\documentclass[preprint]{article}

\usepackage{neurips_2025}

\usepackage[utf8]{inputenc} 
\usepackage[T1]{fontenc}    
\usepackage{hyperref}       
\usepackage{url}            
\usepackage{booktabs}       
\usepackage{amsfonts}       
\usepackage{nicefrac}       
\usepackage{microtype}      

\usepackage{array}

\usepackage{booktabs}
\usepackage{csvsimple}
\usepackage{multirow}
\usepackage{makecell}
\makeatother

\usepackage{balance}

\usepackage{tikz}
\usepackage{pgfplots}
\pgfplotsset{width=8cm,compat=1.13}
\usepackage{arydshln} 
\usetikzlibrary{arrows,shapes,shadows,snakes,positioning,automata,backgrounds,patterns}

\usepackage{graphicx}
\usepackage{subfigure}

\usepackage{caption}
\usepackage{subcaption}

\usepackage{algorithm, algorithmic}

\usepackage{enumitem}

\usepackage{amsmath}
\usepackage{bbm}

\usepackage{amsopn}

\usepackage[most]{tcolorbox}
\usepackage{xcolor}
\usepackage{color}

\definecolor{mycolor51}{RGB}{84,89,105}    
\definecolor{mycolor52}{RGB}{164,117,125}  
\definecolor{mycolor53}{RGB}{231,152,124} 
\definecolor{mycolor54}{RGB}{139,145,182}   
\definecolor{mycolor55}{RGB}{119,113,164} 

\definecolor{mycolor55light}{RGB}{171,164,195}

\definecolor{mycolor41}{RGB}{54,80,131}
\definecolor{mycolor42}{RGB}{183,131,175}
\definecolor{mycolor43}{RGB}{245,166,115}
\definecolor{mycolor44}{RGB}{252,219,114}

\definecolor{mycolor42light}{RGB}{206,178,203}
\definecolor{mycolor42light2}{RGB}{222,204,221}
\definecolor{mycolor43light}{RGB}{245,201,169}
\definecolor{mycolor44light}{RGB}{249,230,175}

\definecolor{mycolor61}{RGB}{17,50,93}
\definecolor{mycolor62}{RGB}{54,80,131}
\definecolor{mycolor63}{RGB}{115,107,157}
\definecolor{mycolor64}{RGB}{183,131,175}
\definecolor{mycolor65}{RGB}{245,166,115}
\definecolor{mycolor66}{RGB}{252,219,114}

\definecolor{mycolor31}{RGB}{189,142,192}
\definecolor{mycolor32}{RGB}{111,128,190}
\definecolor{mycolor33}{RGB}{244,126,98}

\definecolor{mycolor01}{RGB}{103,0,13}
\definecolor{mycolor02}{RGB}{165,15,21}
\definecolor{mycolor03}{RGB}{203,24,29}
\definecolor{mycolor04}{RGB}{239,59,44}
\definecolor{mycolor05}{RGB}{251,106,74}
\definecolor{mycolor06}{RGB}{252,146,114}
\definecolor{mycolor07}{RGB}{252,187,161}
\definecolor{mycolor08}{RGB}{254,224,210}

\definecolor{mycolor09}{RGB}{222,235,247}
\definecolor{mycolor10}{RGB}{198,219,239}
\definecolor{mycolor11}{RGB}{158,202,225}
\definecolor{mycolor12}{RGB}{107,174,214}
\definecolor{mycolor13}{RGB}{66,146,198}
\definecolor{mycolor14}{RGB}{33,113,181}
\definecolor{mycolor15}{RGB}{8,81,156}
\definecolor{mycolor16}{RGB}{8,48,107}

\newcommand{\re}[1]{{\color{red}#1}}

\newcommand{\eg}{{\emph{e.g.,}\ }}
\newcommand{\ie}{{\emph{i.e.,}\ }}

\newtcolorbox{CodeBox}[2][]{%
  title=#2,
  colback=gray!5,
  colframe=gray!75!black,
  fonttitle=\bfseries,
  before=\par\smallskip\centering,
  after=\par\smallskip,
  #1}

\newcommand{\name}[0]{POI-QA}
\newcommand{\datalink}[0]{\href{https://www.kaggle.com/ds/7394666}{https://www.kaggle.com/ds/7394666}}
\newcommand{\codelink}[0]{\href{https://github.com/hahahenha/POI-QA}{https://github.com/hahahenha/POI-QA}}
\title{A Dataset for Spatiotemporal-Sensitive\\POI Question Answering}

\author{%
  Xiao Han*$^1$, 
  Dayan Pan*$^1$, 
  Xuyuan Hu$^2$, 
  Zhaolin Deng$^2$, \\
  \textbf{Xiangjie Kong}$^2$, 
  \textbf{Guojiang Shen}$^2$,
  \textbf{Xiangyu Zhao}$^1$  \\
   $^1$Department of Data Science, City University of Hong Kong, Hong Kong SAR, China \\
  $^2$Department of Computer Science, Zhejiang University of Technology, Hangzhou, China \\
  \texttt{hahahenha@gmail.com}, 
  \texttt{dayanpan2-c@my.cityu.edu.hk},  \\
  \texttt{\{huxuyuan, 211123120032, xjkong, gjshen1975\}@zjut.edu.cn},\\
  \texttt{xianzhao@cityu.edu.hk}
}

\begin{document}

\maketitle

\begin{abstract}
Spatiotemporal relationships are critical in data science, as many prediction and reasoning tasks require analysis across both spatial and temporal dimensions—for instance, navigating an unfamiliar city involves planning itineraries that sequence locations and timing cultural experiences.
However, existing Question-Answering (QA) datasets lack sufficient spatiotemporal-sensitive questions, making them inadequate benchmarks for evaluating models' spatiotemporal reasoning capabilities.
To address this gap, we introduce \name, a novel spatiotemporal-sensitive QA dataset centered on Point of Interest (POI), constructed through three key steps: mining and aligning open-source vehicle trajectory data from GAIA with high-precision geographic POI data, rigorous manual validation of noisy spatiotemporal facts, and generating bilingual (Chinese/English) QA pairs that reflect human-understandable spatiotemporal reasoning tasks.
Our dataset challenges models to parse complex spatiotemporal dependencies, and evaluations of state-of-the-art multilingual LLMs (\emph{e.g.,} Qwen2.5-7B, Llama3.1-8B) reveal stark limitations: even the top-performing model (Qwen2.5-7B fine-tuned with RAG+LoRA) achieves a top 10 Hit Ratio (HR@10) of only 0.41 on the easiest task, far below human performance at 0.56.
This underscores persistent weaknesses in LLMs’ ability to perform consistent spatiotemporal reasoning, while highlighting \name\ as a robust benchmark to advance algorithms sensitive to spatiotemporal dynamics. The dataset is publicly available at \datalink.
\end{abstract}

\section{Introduction}


Spatiotemporal reasoning plays a pivotal role in a wide range of prediction and decision-making tasks that require sensitivity to both spatial and temporal contexts.
This capability depends heavily on spatiotemporal information, which encompasses spatial data, such as geographic locations, and temporal data like time of day or sequential time-based patterns.
As a result, spatiotemporal reasoning has become an essential focus in recent research across domains including mobility analysis, personalized recommendation, and spatiotemporal prediction tasks~ \cite{wan2023spatio,deng2023spatio,wang2021spatio}.
The integration of spatiotemporal reasoning into decision-making processes is not confined to technological applications but is also deeply embedded in the daily routines and choices of individuals~\cite{mateos2025systematic}.
For instance, when planning a journey, travelers often consider factors such as the geographical proximity of restaurants offering local specialties and the time required to reach these establishments.
This example underscores how both spatial and temporal elements are crucial for making informed decisions.
Among the domains where spatiotemporal reasoning is essential, Point of Interest (POI) recommendation stands out as a representative and challenging example. To effectively identify appropriate POIs, models must possess robust spatiotemporal reasoning capabilities. These capabilities enable models to analyze historical user behavior patterns, predict future preferences, and recommend POIs that align with users' interests while accounting for constraints like time and location.
In essence, the ability to reason about space and time is fundamental for developing intelligent recommendation systems that cater to diverse user needs and preferences~\cite{yu2024survey}.

In this paper, we focus on addressing the spatiotemporal challenges of POI prediction with precision and rigor.
We formally define POI prediction at travel destinations as spatiotemporal questions based on the following four criteria:
\textbf{i) Spatiotemporal Presence:} The question contains both a timestamp, [time], and a geolocation, [place], such as ``Tuesday evening'' and ``221B Baker Street'';
\textbf{ii) Spatiotemporal Context Sensitivity:} Answers to similar questions may vary depending on differences in time and/or location, \ie altering the [time] or [place] can result in different answers.
\textbf{iii) Spatiotemporal Knowledge Reasoning:} Such questions require broad POI data coverage and the ability to perform spatiotemporal reasoning.
\textbf{iv) Human-Readable Answer:} The answer should align with effective human-computer interaction principles, such as providing the POI name along with a specific address rather than raw latitude and longitude coordinates.
We found that, despite their ubiquity, spatiotemporal-sensitive questions are under-studied in existing POI QA datasets. 
For example, SubjQA~\cite{bjerva2020subjqa} focuses on attribute-oriented questions derived from POI reviews, requiring only semantic knowledge and lacking spatial or temporal information. MapQA~\cite{li2025mapqa} supports geographic queries but omits any temporal context. TourismQA~\cite{contractor2021answering}, although built from tourism reviews and containing questions related to time or place, lacks the ability to perform spatiotemporal reasoning.
All of these datasets do not consider spatiotemporal-sensitive issues as specified in criterion ii).

\begin{figure}
    \centering
    \includegraphics[width=0.95\linewidth]{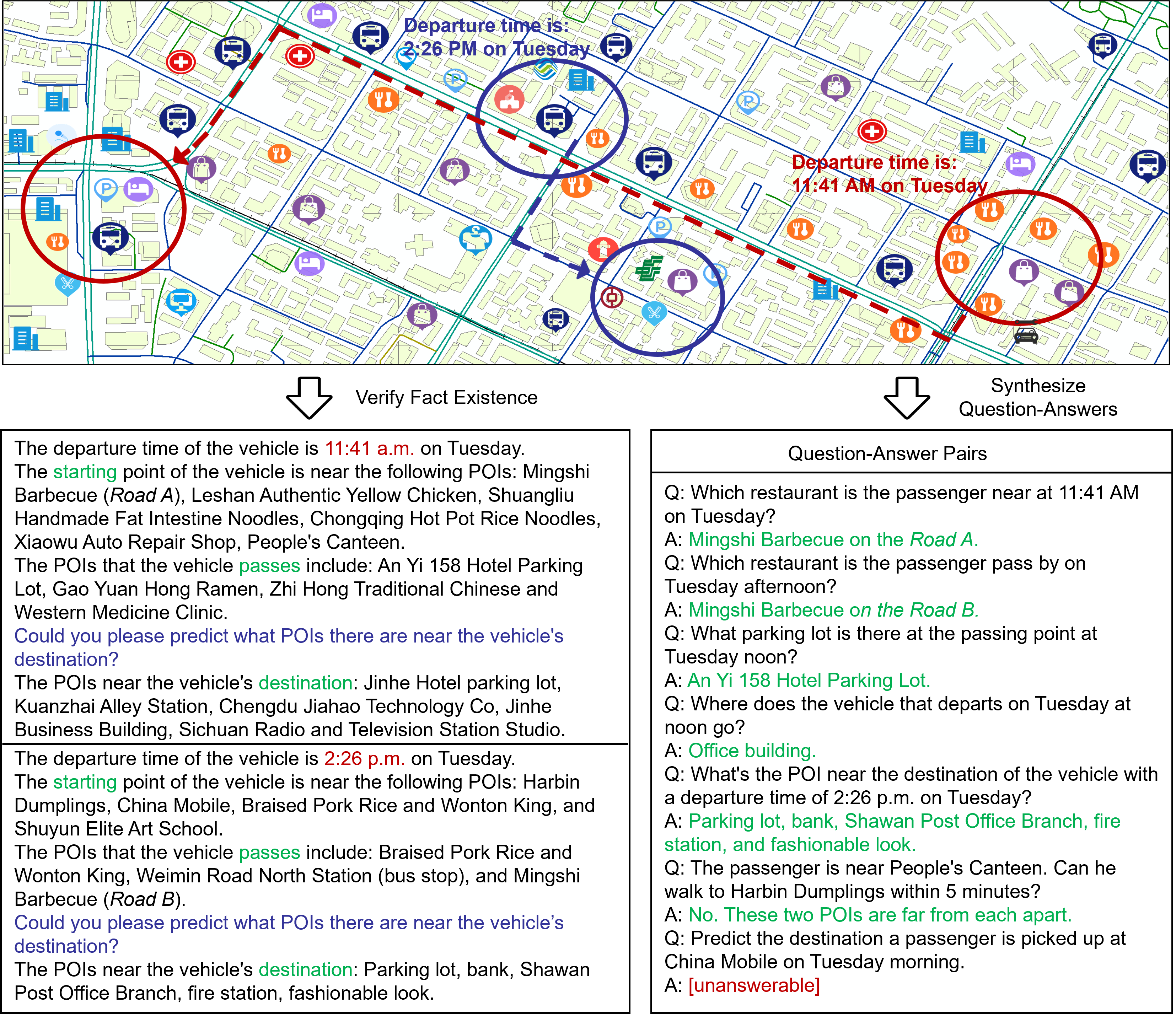}
    \caption{A toy example of spatiotemporal sensitive questions.}
    \label{fig:illustration}
\end{figure}

One of the datasets closest to ours is Foursquare\footnote{https://opensource.foursquare.com/os-places/}, which provides a large amount of POI location information worldwide, along with a large number of user check-in data with timestamps.
However, question samples extracted from the above-mentioned dataset fail to meet criteria ii), iii), and iv).
Furthermore, the spatiotemporal information in the Foursquare dataset is relatively sparse and fragmented, as many users check in at different POIs on the platform with gaps of several days.
Therefore, we propose to construct our own dataset, called \name. We first identify spatiotemporal-evolving relationships from both GAIA trajectory data\footnote{https://outreach.didichuxing.com/} and POI information around those real-time trajectories.
Then, a massive number of human workers are employed to annotate the POIs surrounding each GPS point in every trajectory, especially focusing on double-checking the POIs near pick-up and drop-off locations.
Finally, we created bilingual datasets (in Simplified Chinese and English) with multiple levels of granularity, corresponding to different levels of question difficulty. These levels include POI name, POI subcategory, POI medium category, and POI major category. Each level contains over 5,000,000 question-answer pairs, covering about 400,000 distinct POI locations and 30 consecutive days of vehicle trajectory data.
Using POI names as labels in QA pairs is more challenging, as it requires more spatiotemporal reasoning and natural language understanding compared to other classification tasks.
Figure~\ref{fig:illustration} shows two trajectories and their corresponding QA examples from the \name\ dataset, constructed using both trajectory facts and synthesized contextual information. Although both vehicles depart at similar times on Tuesday, the spatial variation in their departure points leads to different routes and destination contexts. This example highlights the strong spatiotemporal sensitivity of our dataset, where even slight spatial shifts under similar temporal conditions can significantly impact the question context, requiring models to perform spatiotemporal reasoning.
3The challenges posed by our dataset fall into three folds:
\begin{itemize}[leftmargin=*]
    \item \textbf{Geographic Knowledge Processing}: This involves accurately identifying and categorizing POIs based on their geographic locations. For example, recognizing that a ``McDonald's'' in a bustling city center may have different operating hours compared to one in a quieter suburban area.

    \item \textbf{Temporal Information Understanding}: This requires the system to understand how temporal factors affect POI availability or relevance. For instance, recognizing that a restaurant may be open for dinner on weekdays but closed on weekends.

    \item \textbf{Spatiotemporal Reasoning}: This involves combining both geographic and temporal information to provide accurate predictions. For example, recognizing that a user asking about the best places to eat near their home at 8pm is likely looking for a restaurant that is still open and close to home.

\end{itemize}

We evaluate the performance of different state-of-the-art open-source Large Language Models (LLMs) on \name\ across all levels of granularity and observe that the average HR@10 drops from 0.39 on the coarse-grained ``POI Major Category'' task to 0.21 on the fine-grained ``POI Subcategory'' task, indicating that current models struggle with spatiotemporal understanding and reasoning.
In contrast, human performance on the POI Subcategory task reaches an HR@10 of 0.57, highlighting a substantial gap between existing advanced models and human capabilities.
Therefore, we believe \name\ could serve as a valuable benchmark for studying this problem.

\section{\name\ Dataset}

In this section, we demonstrate the pipeline constructing our dataset, \name.
It consists of three steps:
i) geographic annotation of POIs,
ii) trajectory-based POI mapping, and
iii) spatiotemporal question-answer pair generation.

\subsection{Geographic Annotation of POIs}


Before POI annotation, the choice of POI locations is critical \cite{tang2022discovering,DBLP:conf/ijcai/LiCLYH21}: In sparsely populated areas, POIs tend to be distributed sparsely as well, and the resulting datasets are usually of low quality.
On a global scale, Chinese cities have the characteristics of high population density and thriving regional economic activities \cite{ma2023evolution,yu2017chinese}.
These lead to a large number of POIs and rich types, making such cities particularly suitable for POI annotation.
Therefore, we chose Chengdu, a Chinese city with a population in the tens of millions \cite{dong2022spatiotemporal}, as a suitable location for the dataset.

Although POIs in a city, such as store openings, relocations, or closures, may evolve over time, these dynamic changes are simplified in our constructed dataset to ensure consistency with the time frame of GAIA Data.
To align with this requirement, we first collected 418,854 POI entries from map engines as of the end of 2016.
After rigorous screening, we retained 418,579 POIs that remained stable over the period and excluded 275 POIs that had undergone changes.
The POI annotation process followed four core steps:

\textbf{Data Collection via Map Search Engines}:
We crawled POI data from two major map search engines in mainland China: Baidu Maps\footnote{https://lbsyun.baidu.com/} and Amap\footnote{https://lbs.amap.com/}. To ensure comprehensive coverage, we partitioned Chengdu into a grid system of 500x500 cells, each approximately 300 meters in length and width. For each grid, we retrieved and queried nearby POIs of the center point from the search engines.

\textbf{Data Cleaning and Standardization}:
Duplicate entries from the search engine results were removed. Subsequently, we standardized the geographic coordinates of each POI to the WGS84 coordinate system to ensure uniformity \cite{wang2024environmental}.

\textbf{Coordinate Validation and Error Thresholds}:
We calculated the coordinate discrepancy between the same POI across platforms. POIs with a coordinate difference of <1e-4 were retained and recorded. For discrepancies between 1e-4 and 1e-3, a manual review process was conducted to verify and retain valid POIs. Those with errors exceeding 1e-3 were excluded due to potential inaccuracies.

\textbf{Hierarchical Categorization}: In order to describe the POI more clearly, we manually marked all the collected POIs again.
Each POI point has 3 category labels: major category, medium category and subcategory.
For the entire POI dataset, we have divided it into 19 major categories, 122 medium categories and 959 subcategories.
For more details, please refer to Appendix \ref{app:Dataset}.

This systematic approach ensured the reliability and temporal consistency of the POI dataset in alignment with GAIA Data’s requirements.

\subsection{Trajectory-based POI Mapping}

The POI mapping takes three steps: mining spatiotemporal-evolving travel targets from GAIA data, aligning geographic information with POIs, and human verification.

\textbf{Mining Spatiotemporal-evolving Travel Targets from GAIA Data}: We first utilize existing vehicle location records from GAIA Data to identify trajectories with distinctive spatiotemporal migration patterns.
Subsequently, we employ this data to mine trips that exhibit temporal and spatial evolution.
For instance, the vehicle ID ``6c8a8d17e6bbe4cd2fcdb4991b52725e'' in the GAIA Data produces various trip patterns: some travel directly from entertainment venues via main roads to nearby residential areas during weekday evenings, while others divert from community gates to nearby educational institutions on holiday mornings.
These behaviors reflect clear spatiotemporal orientations, such as individuals returning home after nightlife activities or students attending weekend cram schools.
By screening and filtering vehicle trajectory records with discernible objectives, we successfully extracted over 6 million trajectories characterized by prominent spatiotemporal migration patterns.
These trajectories are formatted as: ``carID, timestamp and the location at the pickup point, the positioning sequence during the trip, and the drop-off location.''

\textbf{Aligning Geographic Information with POIs}: 
In the task of predicting points of interest at travel destinations, it is essential to map POIs along trajectories, particularly focusing on those near the start and end points.
This approach addresses the need to avoid private information found in order details or GPS sequences.
Our objectives include anonymization and associating POIs during data processing.
The process involves four key steps: 
i) downsampling the trajectory by retaining positioning information at critical intersections and congestion points while eliminating redundancies;
ii) matching all POIs within a 100-meter radius of start and end points, listed from nearest to farthest;
iii) using the closest POI for journey positioning points to obscure exact paths;
and iv) simplifying timestamps to day of the week and hour.
Each track record is then formatted as: ``anonymous carID, timestamp, POIs near pickup location, POIs during trip, POIs near drop-off location.'' This method ensures privacy while maintaining data utility for effective destination prediction.

\textbf{Human Verification}:
In the prior step, automated programs generate noisy data in batches. The primary sources of errors include: 
i) anomalies and drifts in trajectory points within GAIA data; 
and ii) start or end points situated in city suburbs with low POI coverage, leading to unclear descriptions of trajectory endpoints.
To address these issues, we employ manual verification by hiring workers.
This process involves the following measures:
i) Display the start (end) point of the trajectory alongside nearby POIs (from nearest to farthest, shaded from dark to light) on a single map. Identify and mark records with missing or problematic information, correcting POI details if manually matched.
ii) Visualize downsampled trajectories directly within the road network. Identify and mark trajectories with obvious anomalies or discontinuities, rectifying waypoints as needed.
iii) Assign each trajectory record to at least five different workers for evaluation. If a record is flagged by more than 60\% of evaluators, it is either deleted or adjusted according to the majority opinion.
It ensures data accuracy and reliability through systematic manual verification.

\subsection{Spatiotemporal Question-Answer Pair Generation}

Once we have the precise trajectory-POI matching records, the next step involves generating question-answer pairs that exhibit spatiotemporal correlation.

\begin{table}[ht]
\centering
\footnotesize
\caption{The dataset statistics.}
\label{tab:difficulty}
\begin{tabular}{cccc}
\toprule
\multicolumn{1}{c}{\bfseries{Type}} & \multicolumn{1}{c}{\bfseries{Difficulty}} & \multicolumn{1}{c}{\makecell{\bfseries{Label }\\ \bfseries{Categories}}} & \multicolumn{1}{c}{\bfseries{Specifier}}\\
\midrule
Major Category Classification & Easy & 19 & \makecell{POIs at travel destination are: \\$[$ \re{Lifestyle Services}, \re{Shopping Service}, ...$]$} \\
&&&\\
Medium Category Classification & Medium & 122 & \makecell{POIs at travel destination are: \\$[$\re{Beauty Salon}, \re{Supermarket}, ...$]$} \\
&&&\\
Subcategory Classification & Hard & 959 & \makecell{POIs at travel destination are: \\$[$\re{Plastic Surgery | Healthcare Services}, \\\re{Hui Kang Supermarket}, \\\re{Wanning Supermarket}, ...$]$} \\
&&&\\
POI Name Generation & Very Hard & 400K+
        & \makecell{POIs at travel destination are: \\$[$\re{tai shi xing cai yi xue mei rong}\\\re{(No. 75 Fuqiang Street)}, \re{Wanning}\\ \re{(cheng du fu li guang chang)}, ...$]$} \\
\bottomrule
\end{tabular}
\end{table}

\textbf{Main QA Dataset}:
Our dataset consists of two components. The first part contains POI information, describing the locations and spatial relationships of various POIs. The second part is our main dataset, specifically designed for predicting POIs at travel destinations. Both datasets are generated using templates. Since the data originates from China, we provide both simplified Chinese and English versions to support multilingual model training.

\begin{figure}
    \centering
    \includegraphics[width=0.85\linewidth]{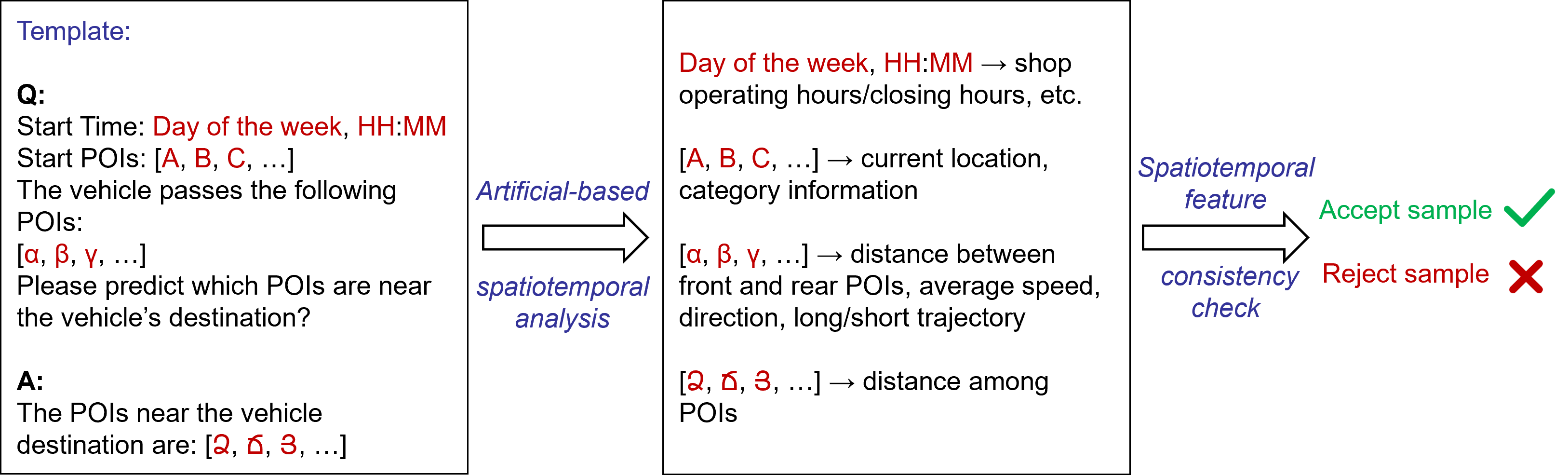}
    \caption{QA sample synthesizing.}
    \label{fig:QA_sample_synthesizing}
\end{figure}

The synthesizing procedure is described in Figure~\ref{fig:QA_sample_synthesizing}.
As shown in Figure~\ref{fig:QA_sample_synthesizing}, we use '<>' to represent the POI name.
Since the English translation of most POIs has no specific meaning, we use the three phrases in '()' to represent the major category, medium category, and subcategory of the POI.
In order to be closer to life and easier for people to understand directly, we also use both addresses in natural language and longitude-latitude coordinates to describe the geographical location of the POI.
Finally, for each POI, we list the nearby POIs and the distances from these POIs to the current POI in the form of an array from near to far.
For the POI prediction sample, we take the POI information near the starting point of the vehicle trajectory and the waypoint as the problem, and take the POI near the end of the vehicle trajectory as the predicted label.
The predicted label is a list form represented by '[]'.
Each record in the list is a POI point, including the POI name and its corresponding three categories.
Therefore, we can use this dataset for two major tasks: classification task and generation task, as shown in Table~\ref{tab:difficulty}.
For the classification task, we hope to build a model that can determine the classification category (major category, medium category, and subcategory) of the POI near the destination; for the generation task, we hope to build a model that can directly output the name of the POI near the destination.
The difficulty of these four tasks increases in turn, and their comprehensive data are shown in Table~\ref{tab:difficulty}.
The license information of the dataset is listed in Appendix~\ref{app:accessibility}.

\textbf{Quality Control}:
In order to obtain a high-quality dataset, we performed very detailed quality control during the collection process. In the interface, we highlight the annotated POIs and timestamps with special fonts to help annotators identify them. We assign each sample to multiple workers at the same time, and let them score the data quality without knowing each other. If the negative score is higher than 60\%, the sample will be removed. In the final verification step, about 20\% of the records were modified, and we finally obtained 5,417,335 high-quality data samples.

\section{Models}

In this section, we first present the formal problem definition for POI prediction at travel destinations.
We then introduce the models used to evaluate the proposed dataset.

\subsection{Learning Problem}

Here we formally define the problem setup.
The model is given a set of POI information $D_{poi} = \{ poi_1, \cdots, poi_N \}$, and questions $Q = \{ q_1, \cdots, q_M \}$, where each POI information $poi_i, i \in [N]$ and question $q_j, j \in [M]$ is a textual sequence of fewer than 8,000 tokens.
The model must possess the following capabilities:
i) Semantic Understanding: Accurately interpret user queries to identify intent and relevant context.
ii) Information Retrieval: Efficiently search through $D_{poi}$ to extract pertinent POI data based on query requirements.
iii) Spatiotemporal Analysis: Incorporate location and time-based constraints to effectively filter and rank candidate POIs. 
iv) Human-Computer Interaction: Generate responses that are not only accurate but also presented in a user-friendly manner, ensuring clarity and relevance.
The model's objective is to generate a response string $\hat{A}$ that accurately answers the query by leveraging these capabilities. This involves selecting the most appropriate POI(s) from $D_{poi}$ based on the query's context and constraints, while maintaining a balance between precision and user experience.
The approach integrates natural language processing techniques with spatiotemporal reasoning to achieve robust performance across diverse scenarios.
\subsection{Pre-trained LLMs with SFT and RAG}

To cope with the existing challenges, especially the four capabilities mentioned in the previous paragraph, we adopt two open-source LLMs as base-models: Llama3.1~\cite{grattafiori2024llama} and Qwen2.5~\cite{yang2024qwen2}, which are known to achieve state-of-the-art performance on a wide range of open-world QA tasks (\eg Natural Question~\cite{kwiatkowski2019natural}, TriviaQA~\cite{joshi2017triviaqa}, and WikiQA~\cite{yang2015wikiqa}).

Llama3.1 and Qwen2.5 are both built with transformer-based decoder architectures with support for a 128K context length.
Llama3.1 introduces Group Query Attention and follows a pretraining pipeline consisting of reward modeling, supervised fine-tuning (SFT), and direct preference optimization (DPO), while Qwen2.5 adopts a two-stage pretraining strategy with RoPE adjusted base frequency (ABF) technology and enhanced Chinese language support.
Appendix~\ref{app:basemodel} provides a detailed description of their design and training process.

Beyond evaluating model performance on \name\ in a zero-shot setting, we also employ Low-Rank Adaptation (LoRA) fine-tuning \cite{hu2022lora} and Retrieval-Augmented Generation (RAG) \cite{lewis2020retrieval} methods for further assessment. More details are provided in Appendix~\ref{app:LoRA} and \ref{app:RAG}.

\section{Experiments}

In this section, we conduct several baseline experiments to better illustrate our proposed dataset.

\subsection{Experimental Setup}
Experiments are conducted using two state-of-the-art LLMs as base models that we mentioned before: Llama3.1-8B and Qwen2.5-7B. 
For the Llama model, we use the English version of the dataset, while the Qwen model uses the Chinese version to generate the best results.
The content of the two versions of the dataset is exactly the same except for languages.
Additionally, we employ one specialized model, Deepseek-r1-32B for fine-grained task decomposition and retrieval results summarization and final generation in the RAG pipeline, as detailed in both the Models section and the Appendix~\ref{app:RAG}. 
We evaluate multiple model variants to analyze the impact of different methods on spatiotemporal reasoning capabilities, including zero-shot, LoRA-based fine-tuning, retrieval-augmented generation (RAG), and a combined LoRA+RAG method.

We utilize a mixed precision training strategy with bf16 to fine-tune all the models using the AdamW optimizer with a learning rate of 1e-4 and a cosine scheduler.
For LoRA-based methods, the rank is set to 16. Models are fine-tuned for 3 epochs, using a batch size of 24 per GPU.
The best model is selected based on validation set performance which is constituted from 10\% of the total dataset.
All training is conducted on NVIDIA A100 with 80G memory running Ubuntu 22.04.

\subsection{Evaluation Metrics}

We evaluate model performance on four answer types: POI name, subcategory, medium category, and major category, covering spatiotemporal reasoning at multiple granularities.
We designed two evaluation settings differing in how the answer space is defined: \textbf{QA for Classification Tasks} and \textbf{Open-world Generative QA}.
For both settings, we report Hit Ratio (HR@$k$) and Normalized Discounted Cumulative Gain (NDCG@$k$) at $k\!\in\!\{5,10,20\}$.
For the generative setting, we additionally compute BLEU-based textual-similarity scores to assess lexical quality.
Detailed metric definitions are provided in Appendix~\ref{app:metrics}.

\subsection{Main Results}
\label{exp:main_results}

Tables \ref{tab:classification_hr}--\ref{tab:generation_results}
summarize the primary results across model variants and metrics for the classification tasks and open-world generative QA tasks, respectively.
Each table reports the performance of the base LLMs, Qwen2.5-7B and Llama3.1-8B, under four experimental configurations: zero-shot, LoRA-based fine-tuning, RAG, and combined RAG+LoRA.

\paragraph{QA for Classification tasks.}
As shown in both Tables~\ref{tab:classification_hr} and \ref{tab:classification_ndcg}, zero-shot performance is consistently low, confirming that spatiotemporal reasoning remains challenging for out-of-the-box LLMs.
LoRA and RAG can both enhance model performance.
Taking $k=10$ as an example,
LoRA contributes an improvement of 0.05 and 0.09 in HR@10 on average for Llama and Qwe, whereas RAG, through the integration of external spatiotemporal knowledge, achieves a sightly larger gain of 0.06 and 0.13.
When combined, RAG + LoRA obtains the best result, outperforming the zero-shot baseline 2.5 and 3.9 times on HR@10 and NDCG@10, respectively.

\begin{table}[ht]
\centering
\caption{Results for classification tasks. We report HR@\{5,10,20\} for each model variant.}
\label{tab:classification_hr}
\small
\resizebox{1.\linewidth}{!}{
    \begin{tabular}{l|ccc|ccc|ccc}
    \toprule
    \multirow{2}{*}{\textbf{Model}}
    & \multicolumn{3}{c|}{\textbf{Major Category}}
    & \multicolumn{3}{c|}{\textbf{Medium Category}}
    & \multicolumn{3}{c}{\textbf{Subcategory}} \\
    \cmidrule(lr){2-4} \cmidrule(lr){5-7} \cmidrule(lr){8-10}
    & \textbf{{\color{white}{H}}HR@5{\color{white}{H}}} & \textbf{{\color{white}{H}}HR@10{\color{white}{H}}} & \textbf{{\color{white}{H}}HR@20{\color{white}{H}}}
    & \textbf{{\color{white}{H}}HR@5{\color{white}{H}}} & \textbf{{\color{white}{H}}HR@10{\color{white}{H}}} & \textbf{{\color{white}{H}}HR@20{\color{white}{H}}}
    & \textbf{{\color{white}{H}}HR@5{\color{white}{H}}} & \textbf{{\color{white}{H}}HR@10{\color{white}{H}}} & \textbf{{\color{white}{H}}HR@20{\color{white}{H}}}\\
    \midrule 
    Llama3.1-8B (zero-shot)
    & 0.0664 & 0.1001 & 0.0917
    & 0.0281 & 0.0481 & 0.0695
    & 0.0222 & 0.0350 & 0.0372 \\
    Qwen2.5-7B (zero-shot)
    & 0.1017 & 0.1775 & 0.1650
    & 0.0451 & 0.0784 & 0.0814
    & 0.0263 & 0.0467 & 0.0673 \\
    \midrule
    Llama3.1-8B (LoRA)
    & 0.1239 & 0.1880 & 0.2067
    & 0.0590 & 0.1041 & 0.1241
    & 0.0445 & 0.0687 & 0.0797 \\
    Qwen2.5-7B (LoRA)
    & 0.1950 & 0.3222 & 0.3509
    & 0.1004 & 0.1627 & 0.1871
    & 0.0611 & 0.1062 & 0.1250 \\
    \midrule
    Llama3.1-8B (RAG)
    & 0.1237 & 0.1770 & 0.2089
    & 0.0593 & 0.1155 & 0.1328
    & 0.0461 & 0.0721 & 0.0848 \\
    Qwen2.5-7B (RAG)
    & 0.2099 & \underline{0.3821} & 0.3815
    & 0.0967 & 0.1876 & 0.2008
    & 0.0650 & 0.1107 & 0.1218 \\
    \midrule
    Llama3.1-8B (RAG+LoRA)
    & \underline{0.2189} & 0.3784 & \underline{0.4356}
    & \underline{0.1736} & \underline{0.2966} & \underline{0.3379}
    & \underline{0.1092} & \underline{0.2009} & \underline{0.2324} \\
    Qwen2.5-7B (RAG+LoRA)
    & \textbf{0.2339} & \textbf{0.4062} & \textbf{0.4698}
    & \textbf{0.1812} & \textbf{0.2987} & \textbf{0.3577}
    & \textbf{0.1288} & \textbf{0.2185} & \textbf{0.2586} \\
    \bottomrule
    \end{tabular}
}

\small{
Bold and underlined indicates statistically significant improvement \\(\ie using a two-sided t-test with $p<0.05$) over the best baseline.
}
\end{table}

\begin{table}[ht]
\centering
\caption{Results for classification tasks. We report NDCG@\{5,10,20\} for each model variant.}
\label{tab:classification_ndcg}
\small
\resizebox{1.\linewidth}{!}{
    \begin{tabular}{l|ccc|ccc|ccc}
    \toprule
    \multirow{2}{*}{\textbf{Model}}
    & \multicolumn{3}{c|}{\textbf{Major Category}}
    & \multicolumn{3}{c|}{\textbf{Medium Category}}
    & \multicolumn{3}{c}{\textbf{Subcategory}} \\
    \cmidrule(lr){2-4} \cmidrule(lr){5-7} \cmidrule(lr){8-10}
    & \textbf{NDCG@5} & \textbf{NDCG@10} & \textbf{NDCG@20}
    & \textbf{NDCG@5} & \textbf{NDCG@10} & \textbf{NDCG@20}
    & \textbf{NDCG@5} & \textbf{NDCG@10} & \textbf{NDCG@20}\\

    \midrule 
    Llama3.1-8B (zero-shot)
    & 0.1073 & 0.1841 & 0.2150
    & 0.0617 & 0.1241 & 0.1380
    & 0.0631 & 0.0842 & 0.1141 \\
    Qwen2.5-7B (zero-shot)
    & 0.1778 & 0.3130 & 0.3521
    & 03.1047 & 0.1736 & 0.2369
    & 0.0910 & 0.1319 & 0.1642 \\
    \midrule
    Llama3.1-8B (LoRA)
    & 0.2085 & 0.3448 & 0.3948
    & 0.1284 & 0.2268 & 0.2646
    & 0.1182 & 0.1959 & 0.2247 \\
    Qwen2.5-7B (LoRA)
    & 0.3555 & 0.5694 & 0.6976
    & 0.1968 & 0.3479 & 0.4270
    & 0.1898 & 0.2804 & 0.3241 \\
    \midrule
    Llama3.1-8B (RAG)
    & 0.2436 & 0.3911 & 0.4029
    & 0.1319 & 0.2530 & 0.2857
    & 0.1304 & 0.2075 & 0.2245 \\
    Qwen2.5-7B (RAG)
    & 0.3550 & 0.6315 & 0.6790
    & 0.2121 & 0.3655 & 0.4646
    & 0.1879 & 0.2808 & 0.3250 \\
    \midrule
    Llama3.1-8B (RAG+LoRA)
    & \underline{0.4722} & \underline{0.6940} & \underline{0.7363}
    & \underline{0.3512} & \underline{0.6464} & \underline{0.7485}
    & \underline{0.3512} & \underline{0.5729} & \underline{0.6595} \\
    Qwen2.5-7B (RAG+LoRA)
    & \textbf{0.4615} & \textbf{0.7179} & \textbf{0.8307}
    & \textbf{0.3699} & \textbf{0.6388} & \textbf{0.7118}
    & \textbf{0.3143} & \textbf{0.5767} & \textbf{0.6822} \\
    \bottomrule
    \end{tabular}
}

\small{
Bold and underlined indicates statistically significant improvement \\(\ie using a two-sided t-test with $p<0.05$) over the best baseline.
}
\end{table}

\begin{table}[ht]
\centering
\caption{Open-world Generative QA results.  
Besides HR@\{5,10,20\} and NDCG@\{5,10,20\}, we include BERTScore\textsubscript{F1} (“BLEUScore” column) to measure lexical similarity.}
\label{tab:generation_results}
\small
\resizebox{1.\linewidth}{!}{
    \begin{tabular}{l|ccc|ccc|c}
    \toprule
    \multirow{2}{*}{\textbf{Model}}
    & \multicolumn{3}{c|}{\textbf{Hit Ratio (Full Match)}}
    & \multicolumn{3}{c|}{\textbf{NDCG  (Full Match)}} 
    &     \multirow{2}{*}{\textbf{BLEUScore}}
    \\
    \cmidrule(lr){2-4} \cmidrule(lr){5-7}
    & \textbf{HR@5} & \textbf{HR@10} & \textbf{HR@20}
    & \textbf{NDCG@5} & \textbf{NDCG@10} & \textbf{NDCG@20} \\
    
    \midrule 
    Llama3.1-8B (zero-shot)
    & 0.0075 & 0.0112 & 0.0146
    & 0.0149 & 0.0244 & 0.0297
    & 0.0332 \\
    Qwen2.5-7B (zero-shot)
    & 0.0119 & 0.0199 & 0.0234
    & 0.0213 & 0.0390 & 0.0442
    & 0.0254 \\
    \midrule
    Llama3.1-8B (LoRA)
    & 0.0144 & 0.0241 & 0.0282
    & 0.0320 & 0.0512 & 0.0589
    & 0.2941 \\
    Qwen2.5-7B (LoRA)
    & 0.0220 & 0.0394 &0.0459
    & 0.0464 & 0.0798 & 0.0940
    & 0.3082 \\
    \midrule
    Llama3.1-8B (RAG)
    & 0.0142 & 0.0232 & 0.0294
    & 0.0338 & 0.0537 & 0.0640
    & 0.4125 \\
    Qwen2.5-7B (RAG)
    & 0.0226 & 0.0441 & 0.0496
    & 0.0484 & 0.0850 & 0.1048
    & 0.5321 \\
    \midrule
    Llama3.1-8B (RAG+LoRA)
    & \underline{0.0331} & \underline{0.0584} & \underline{0.0690}
    & \underline{0.0725} & \underline{0.1276} & \textbf{0.1509}
    & \underline{0.7729} \\
    Qwen2.5-7B (RAG+LoRA)
    & \textbf{0.0394} & \textbf{0.0616} & \textbf{0.0714}
    & \textbf{0.0770} & \textbf{0.1289} & \underline{0.1508}
    & \textbf{0.7911} \\
    \bottomrule
    \end{tabular}
}

\small{
Bold and underlined indicates statistically significant improvement \\(\ie using a two-sided t-test with $p<0.05$) over the best baseline.
}
\end{table}

\paragraph{Open-world Generative QA.}
This task poses a greater challenge, as models are required not only to reason over complex spatiotemporal constraints but also to generate accurately formatted POI names.
Taking $k=10$ as an instance,
in the zero-shot setting, HR@10 drops to 0.0075 for Llama and 0.0119 for Qwen, and even the best-performing configuration, RAG combined with LoRA, achieves only 0.06 for HR@10 on average and 0.1283 for NDCG@10 on average.

Despite the difficulty, both LoRA and RAG contribute positively.
LoRA increases HR@10 by almost 100\%, RAG provides an additional improvement of about 110\%, and their combination yields a total gain of 6 times than the zero-shot setting.
While the strict ranking metrics remain relatively low, the BLEUScore maintains relatively high when combining with RAG \& LoRA approaches, indicating that the generated outputs are often semantically similar to the label even when they do not match exactly.
This finding highlights the necessity of controlling hallucination and ensuring accurate outputs in generative spatiotemporal QA tasks.
However, the differentiated results also indicate that the proposed dataset requires a more precise spatiotemporal relationship analysis modeling method to improve its accuracy.

\begin{table}[ht]
\centering
\caption{Performance on the human-paraphrased subset of \name.}
\label{tab:human_results}
\small
\resizebox{1.\linewidth}{!}{
    \begin{tabular}{l|ccc|ccc|c}
    \toprule
     \multicolumn{1}{l}{\multirow{2}{*}{\textbf{Task}}} 
    & \multicolumn{3}{c}{\textbf{Hit Ratio}}
    & \multicolumn{3}{c}{\textbf{NDCG}} 
    & \multirow{2}{*}{\textbf{BLEUScore}}
    \\
    \cmidrule(lr){2-4} \cmidrule(lr){5-7}
    & \textbf{HR@5} & \textbf{HR@10} & \textbf{HR@20}
    & \textbf{NDCG@5} & \textbf{NDCG@10} & \textbf{NDCG@20} \\

    \midrule 
    Classification: Major Category
    & 0.3493 & 0.5644 & 0.6701
    & 0.6518 & 0.7774 & 0.8432
    & - \\
    Classification: Medium Category
    & 0.2891 & 0.4150 & 0.4693
    & 0.5119 & 0.6875 & 0.7861
    & - \\
    Classification: Subcategory
    & 0.1833 & 0.3035 & 0.3481
    & 0.4411 & 0.6012 & 0.7140
    & - \\
    \midrule
    Generation:\quad\ POI Names
    & 0.1548 & 0.1611 & 0.1984
    & 0.2096 & 0.2667 & 0.2924
    & 0.8655 \\
    \bottomrule
    \end{tabular}
}

\small{
Bold and underlined indicates statistically significant improvement \\(\ie using a two-sided t-test with $p<0.05$) over the best baseline.
}
\end{table}

\subsection{Human-Paraphrased Results}
\label{exp:human_para}

To assess how well the models generalize to natural user queries, we asked crowd-workers to paraphrase $N_{\text{para}}{=}1{,}000$ questions in \name's test data.
Table~\ref{tab:human_results} reports the results for the zero-shot and the best baseline RAG+LoRA.
Besides we report the result of the model finetuned on RAG+LoRA.
Across the two base LLMs, the performance drop from template to paraphrased questions is quite significant, roughly 70\% on HR on average and 85\% on NDCG on average.

\section{Related Work}

\subsection{POI-related QA}
In recent years, many works have been proposed on POI-related tasks, particularly with the rise of location-based services. 
Early datasets often involved retrieving factual data from structured knowledge bases or user-generated content. 
For instance, 
POIReviewQA~\cite{mai2018poireviewqa} have been proposed to support open-domain search and QA by using Yelp reviews.
Tourism Reviews are also involved in building POI recommendation questions~\cite{contractor2021answering}.
More recently, MapQA~\cite{li2025mapqa} focuses on open-domain QA on geospatial entities and relationships, using geospatial data as the reference.

While these datasets advance POI-related QA by leveraging user reviews and geospatial data, they primary focus on knowledge extraction from static information or direct user preference modeling, rather than systematically evaluating a model's spatiotemporal reasoning capabilities. Thus, we hope our dataset could serve as a complement to the existing POI-related QA research.

\subsection{Spatiotemporal Reasoning}
Spatiotemporal reasoning, which involves understanding and making inferences based on the combined dimensions of space and time, is crucial for many AI applications. In NLP and QA, several efforts have targeted temporal reasoning. 
For example, recent datasets like TempQuestions~\cite{jia2018tempquestions} and the ComplexTempQA~\cite{gruber2024complextempqa} specifically focus on temporal question answering, with the latter tackling complex queries requiring across-time comparison and multi-hop temporal reasoning. On the spatial side, datasets like MapQA~\cite{li2025mapqa} evaluate the performance of geospatial reasoning by using map data directly. 

However, many of these datasets treat temporal and spatial aspects with a primary focus on one or the other. \name~aims to fill this gap by providing QA that explicitly considers spatiotemporal dependency in the context of POI trajectories.

\subsection{Spatiotemporal Foundation LLMs}

LLMs have strong capabilities in general question answering, but there is still much room for spatiotemporal reasoning in specific dynamic real-world scenarios.
Recently, research has increasingly focused on specialized adaptations to improve LLM's spatiotemporal understanding and reasoning.
For instance, the CityGPT~\cite{feng2024citygpt} aims to empower the urban spatial cognition of LLMs by fine-tuning them with a specially constructed instruction dataset, CityInstruction, to introduce urban knowledge and enhance spatial reasoning for city-scale tasks. BIGCity~\cite{yu2024bigcity} proposes a universal spatiotemporal model for a unified analysis of diverse spatiotemporal data types.

Besides, benchmarks like STBench~\cite{li2024stbench} assess LLMs on a range of spatio-temporal tasks, including knowledge comprehension, spatio-temporal reasoning, accurate computation, and downstream applications.
Our \name~highlights the spatiotemporal-sensitive questions for evaluating models' spatiotemporal reasoning.

\section{Conclusion}

In this paper, we explored the importance of spatiotemporal reasoning in real-world tasks.
We highlighted the limitations of existing QA datasets illustrating spatiotemporal-sensitive questions and introduced a novel dataset called \name\ to address these challenges.
This dataset incorporates real-world de-privacy trajectory data and extensive human annotations, providing a comprehensive resource for evaluating spatiotemporal reasoning capabilities.

Our analysis revealed significant performance drops in state-of-the-art models on refined POI prediction tasks, underscoring the need for improved spatiotemporal understanding. With its unique features, including bilingual support and diverse granularities, \name\ serves as a valuable benchmark for advancing research in intelligent recommendation systems. We believe it will play a pivotal role in developing more accurate and context-aware solutions for real-world applications.








\clearpage

\section*{NeurIPS Paper Checklist}

The checklist is designed to encourage best practices for responsible machine learning research, addressing issues of reproducibility, transparency, research ethics, and societal impact. Do not remove the checklist: {\bf The papers not including the checklist will be desk rejected.} The checklist should follow the references and follow the (optional) supplemental material.  The checklist does NOT count towards the page
limit. 

Please read the checklist guidelines carefully for information on how to answer these questions. For each question in the checklist:
\begin{itemize}
    \item You should answer \answerYes{}, \answerNo{}, or \answerNA{}.
    \item \answerNA{} means either that the question is Not Applicable for that particular paper or the relevant information is Not Available.
    \item Please provide a short (1–2 sentence) justification right after your answer (even for NA). 
\end{itemize}

{\bf The checklist answers are an integral part of your paper submission.} They are visible to the reviewers, area chairs, senior area chairs, and ethics reviewers. You will be asked to also include it (after eventual revisions) with the final version of your paper, and its final version will be published with the paper.

The reviewers of your paper will be asked to use the checklist as one of the factors in their evaluation. While "\answerYes{}" is generally preferable to "\answerNo{}", it is perfectly acceptable to answer "\answerNo{}" provided a proper justification is given (e.g., "error bars are not reported because it would be too computationally expensive" or "we were unable to find the license for the dataset we used"). In general, answering "\answerNo{}" or "\answerNA{}" is not grounds for rejection. While the questions are phrased in a binary way, we acknowledge that the true answer is often more nuanced, so please just use your best judgment and write a justification to elaborate. All supporting evidence can appear either in the main paper or the supplemental material, provided in appendix. If you answer \answerYes{} to a question, in the justification please point to the section(s) where related material for the question can be found.

IMPORTANT, please:
\begin{itemize}
    \item {\bf Delete this instruction block, but keep the section heading ``NeurIPS Paper Checklist"},
    \item  {\bf Keep the checklist subsection headings, questions/answers and guidelines below.}
    \item {\bf Do not modify the questions and only use the provided macros for your answers}.
\end{itemize}


\begin{enumerate}

\item {\bf Claims}
    \item[] Question: Do the main claims made in the abstract and introduction accurately reflect the paper's contributions and scope?
    \item[] Answer: \answerYes{} 
    \item[] Justification: 
    The abstract and introduction accurately reflect the paper’s contribution by presenting the development of the POI-QA dataset, and addressing the gap in spatiotemporal-sensitive question answering. We also capture the scope by highlighting the importance of spatiotemporal reasoning and establishing POI-QA as a benchmark for advancing algorithms in this domain.

\item {\bf Limitations}
    \item[] Question: Does the paper discuss the limitations of the work performed by the authors?
    \item[] Answer: \answerYes{} 
    \item[] Justification: We discuss the limitation of the current models dealing with spatiotemporal-sensitive QA dataset in both Introduction, Experiments, and Appendix sections. 

\item {\bf Theory assumptions and proofs}
    \item[] Question: For each theoretical result, does the paper provide the full set of assumptions and a complete (and correct) proof?
    \item[] Answer: \answerNA{} 
    \item[] Justification: This paper mainly proposed a novel spatiotemporal-sensitive QA dataset. 

    \item {\bf Experimental result reproducibility}
    \item[] Question: Does the paper fully disclose all the information needed to reproduce the main experimental results of the paper to the extent that it affects the main claims and/or conclusions of the paper (regardless of whether the code and data are provided or not)?
    \item[] Answer: \answerYes{} 
    \item[] Justification: We published both dataset (\datalink) and the source code (\codelink).

\item {\bf Open access to data and code}
    \item[] Question: Does the paper provide open access to the data and code, with sufficient instructions to faithfully reproduce the main experimental results, as described in supplemental material?
    \item[] Answer: \answerYes{} 
    \item[] Justification: The dataset and code link are listed in the last answer. Besides, we also detailed the experimental settings in both Experiments and Appendix sections.

\item {\bf Experimental setting/details}
    \item[] Question: Does the paper specify all the training and test details (e.g., data splits, hyperparameters, how they were chosen, type of optimizer, etc.) necessary to understand the results?
    \item[] Answer: \answerYes{} 
    \item[] Justification: 
    We provide a detailed settings for experiments.

\item {\bf Experiment statistical significance}
    \item[] Question: Does the paper report error bars suitably and correctly defined or other appropriate information about the statistical significance of the experiments?
    \item[] Answer: \answerYes{} 
    \item[] Justification: We use two-sided t-test with $p < 0.05$.

\item {\bf Experiments compute resources}
    \item[] Question: For each experiment, does the paper provide sufficient information on the computer resources (type of compute workers, memory, time of execution) needed to reproduce the experiments?
    \item[] Answer: \answerYes{} 
    \item[] Justification: We provide the computer resources in the second paragraph of Section 4.1, Experimental Setup.
    
\item {\bf Code of ethics}
    \item[] Question: Does the research conducted in the paper conform, in every respect, with the NeurIPS Code of Ethics \url{https://neurips.cc/public/EthicsGuidelines}?
    \item[] Answer: \answerYes{} 
    \item[] Justification:
    The paper complies with the NeurIPS Code of Ethics.

\item {\bf Broader impacts}
    \item[] Question: Does the paper discuss both potential positive societal impacts and negative societal impacts of the work performed?
    \item[] Answer: \answerYes{} 
    \item[] Justification: 
    We mentioned those in Appendix.
    
\item {\bf Safeguards}
    \item[] Question: Does the paper describe safeguards that have been put in place for responsible release of data or models that have a high risk for misuse (e.g., pretrained language models, image generators, or scraped datasets)?
    \item[] Answer: \answerYes{} 
    \item[] Justification: 
    We conducted a detailed manual review and ensured that the dataset had no privacy issues.

\item {\bf Licenses for existing assets}
    \item[] Question: Are the creators or original owners of assets (e.g., code, data, models), used in the paper, properly credited and are the license and terms of use explicitly mentioned and properly respected?
    \item[] Answer: \answerYes{} 
    \item[] Justification:
    We mentioned them in Appendix.

\item {\bf New assets}
    \item[] Question: Are new assets introduced in the paper well documented and is the documentation provided alongside the assets?
    \item[] Answer: \answerYes{} 
    \item[] Justification: 
    We provide a detailed README of the proposed dataset on the public dataset webpage.

\item {\bf Crowdsourcing and research with human subjects}
    \item[] Question: For crowdsourcing experiments and research with human subjects, does the paper include the full text of instructions given to participants and screenshots, if applicable, as well as details about compensation (if any)? 
    \item[] Answer: \answerNA{} 
    \item[] Justification: None.

\item {\bf Institutional review board (IRB) approvals or equivalent for research with human subjects}
    \item[] Question: Does the paper describe potential risks incurred by study participants, whether such risks were disclosed to the subjects, and whether Institutional Review Board (IRB) approvals (or an equivalent approval/review based on the requirements of your country or institution) were obtained?
    \item[] Answer: \answerNA{} 
    \item[] Justification: None.

\item {\bf Declaration of LLM usage}
    \item[] Question: Does the paper describe the usage of LLMs if it is an important, original, or non-standard component of the core methods in this research? Note that if the LLM is used only for writing, editing, or formatting purposes and does not impact the core methodology, scientific rigorousness, or originality of the research, declaration is not required.
    \item[] Answer: \answerYes{} 
    \item[] Justification: 
    This paper mainly proposes a spatiotemporal sensitive POI dataset and uses LLM fine-tuning to verify the proposed dataset. This work introduces foundation LLMs used and the fine-tuning method in detail.

\end{enumerate}

\clearpage

\appendix

\section{Dataset documentation}
\label{app:Dataset}

We follow datasheets for datasets guideline to document the followinging aspects.

\subsection{Motivation}

\begin{itemize}
    \item What is the purpose this dataset being created? Was there a specific task or gap in existing datasets that it aimed to address?

    The dataset was created to enhance spatiotemporal reasoning in POI prediction at travel destinations. It aimed to address the gap in existing QA models by providing comprehensive data for tasks requiring accurate predictions based on both spatial and temporal information. The specific goal was to improve model performance in understanding and predicting user preferences while accounting for constraints like time and location, ultimately filling a critical gap in datasets that do not adequately represent real-world spatiotemporal dynamics.

    \item What is the source of the dataset, and who create it?

    The data mainly comes from GAIA's open-source 2016 Chengdu online car-hailing trajectory data and the local POI information collected by the team in 2016. The main creator is Xiao Han.
    
\end{itemize}

\subsection{Dataset Overview}

This dataset is designed for spatio-temporal POI prediction tasks.
It has the following file structures:

\begin{itemize}
    \item CN
    \begin{itemize}
        \item POI\_CN.txt
        \item traj\_CN\_train.txt
        \item traj\_CN\_test.txt
        \item traj\_CN\_val.txt
    \end{itemize}
    \item ENG
    \begin{itemize}
        \item POI\_ENG.txt
        \item traj\_ENG\_train.txt
        \item traj\_ENG\_test.txt
        \item traj\_ENG\_val.txt
    \end{itemize}
    \item category
    \begin{itemize}
        \item major\_categoty.csv
        \item medium\_category.csv
        \item subcategory.csv
    \end{itemize}
\end{itemize}

The data has two versions, Simplified Chinese and English, which are stored in the 'CN' and 'ENG' folders, respectively. Each folder has the following files:

\begin{itemize}
    \item The division ratio of the train, val, and test is 8:1:1.

    \item Traj\_XX\_train.csv: Chinese version of the trajectory train data, Sample size: 4,333,868.

    \item Traj\_XX\_test.csv: Chinese version of the trajectory test data, Sample size: 541,733.

    \item Traj\_XX\_val.csv: Chinese version of the trajectory val data, Sample size: 541,734.

    \item POI\_XX.txt: The POI information.
\end{itemize}

Additionally, we developed a Chinese-English vocabulary list for the three-tier classification of POIs, which includes 19 major categories, 122 medium categories, and 959 subcategories. This resource is available in the 'category' folder and can assist scholars in performing classification tasks across different granularities.

\subsection{Uses}

\begin{itemize}
    \item What are the specific tasks that the dataset is designed for?

    QA tasks, especially those requiring spatiotemporal reasoning.

    \item Is there a website or online resource that hosts any of the papers or systems utilizing this dataset?

    No. The dataset, \name, is a novel dataset and we release it at \href{https://www.kaggle.com/ds/7394666}{https://www.kaggle.com/ds/ 7394666}.

    \item What other refined tasks could this dataset be used for?

    POI classification, POI generation, POI retrieval and so on.

    \item What are the potential positive and negative societal impacts?

    The introduction of POI-QA represents a significant advancement in spatiotemporal reasoning within data science, offering potential positive impacts such as enhancing navigation systems with better temporal awareness, improving urban planning by optimizing location-based decisions, and increasing accessibility for multilingual users through its bilingual design. However, it also raises concerns about the potential risk of errors in real-world applications due to current limitations in models' spatiotemporal reasoning capabilities. Thus, while POI-QA holds great promise for advancing AI research and practical applications, careful consideration is needed to address these challenges.
\end{itemize}

\subsection{Distribution}

\begin{itemize}
    \item How will the dataset will be distributed (e.g., tarball on website, API, GitHub)? Does the
dataset have a digital object identifier (DOI)?

Release on Kaggle. DOI: 10.34740/KAGGLE/DS/7394666.

\item Will the dataset be distributed under a copyright or other intellectual property (IP) license,
and/or under applicable terms of use (ToU)? If so, please describe this license and/or ToU,
and provide a link or other access point to, or otherwise reproduce, any relevant licensing
terms or ToU, as well as any fees associated with these restrictions.

Yes. CC BY-NC 4.0 license.

\item Are there any privacy concerns with using this dataset?

The POI data used in this dataset was sourced from city records dating back nearly 10 years (current year: 2025). Since POI information evolves over time, there are no privacy concerns associated with it. Additionally, we removed all GPS details from the vehicle trajectory data and obscured the precise timestamps to ensure that no user privacy information is included in the dataset.

\end{itemize}

\subsection{Accessibility}
\label{app:accessibility}

\begin{itemize}
    \item The link to access the dataset and its metadata: \datalink.

    \item The link of the test code: \codelink.

    \item License: CC BY-NC 4.0. For more details, please refer to \href{https://creativecommons.org/licenses/by-nc/4.0/}{https://creativecommons.org/ licenses/by-nc/4.0/}.
\end{itemize}

\section{Model Design and Training Details}
\label{app:basemodel}
\begin{figure}
    \centering
    \includegraphics[width=0.95\linewidth]{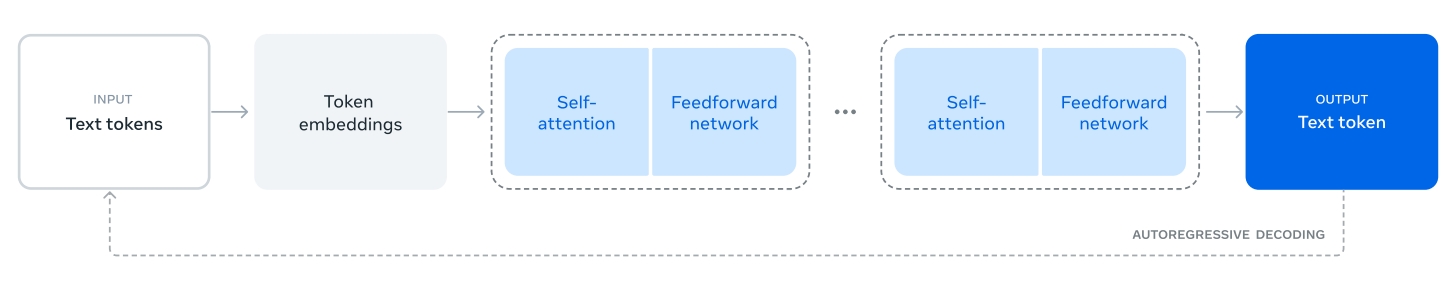}
    \caption{Transformer-based Decoder Structure}
    \label{fig:LLM_structure}
\end{figure}

\begin{figure}
    \centering
    \includegraphics[width=0.95\linewidth]{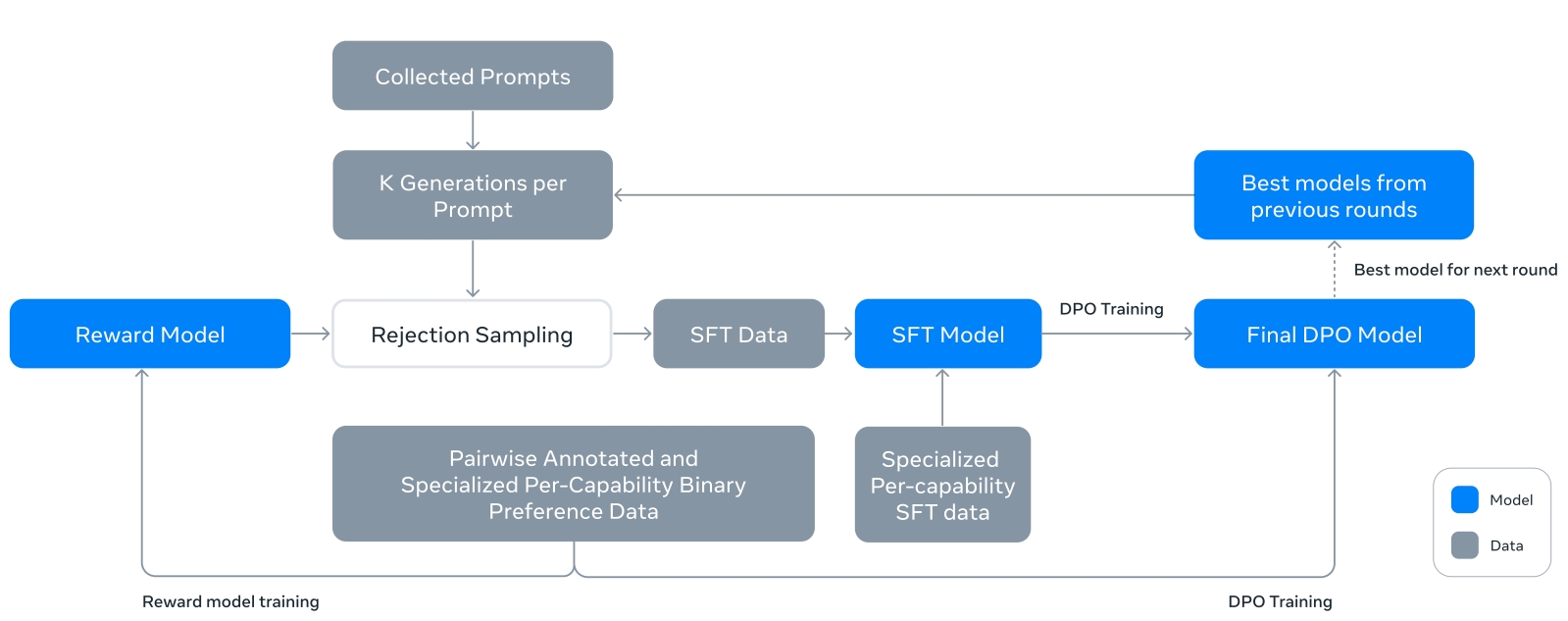}
    \caption{Training Procedure of Llama3.1}
    \label{fig:llama3.1training}
\end{figure}

\textbf{Llama3.1}:
The model employs a Transformer-based decoder architecture (see Figure \ref{fig:LLM_structure}), supporting a context length of 128K. To enhance reasoning efficiency, it incorporates Group Query Attention (GQA). The training process follows a structured approach: initial reward modeling, followed by Supervised Fine-Tuning (SFT), and concluding with Direct Preference Optimization (DPO), as depicted in Figure \ref{fig:llama3.1training}.
This model has achieved state-of-the-art results in trivia-based QA and domain-specific reasoning, demonstrating strong adaptability across various datasets \cite{grattafiori2024llama}.
Its capability to efficiently process lengthy sequences makes it ideal for applications involving detailed documents or complex problem-solving tasks.

\textbf{Qwen2.5}:
The dense architecture of this model mirrors that of LLaMA 3.1, utilizing a transformer-based decoder structure with support for a 128K context length. During pre-training, this model differs by employing a two-stage method that incorporates ABF technology to expand RoPE frequency bases, as opposed to LLaMA's three-stage process. During post-training stage, Qwen2.5 primarily utilizes SFT and multi-stage RL techniques, replacing the DPO methodology used in LLaMA 3.1.
In addition, it offers enhanced support for Chinese corpus compared to its counterpart.

\section{Model Testing with LoRA}
\label{app:LoRA}

In this section, we will detail the LoRA fine-tuning process in our paper.
We use LLaMA-Factory to directly perform LoRA fine-tuning: https://github.com/hiyouga/LLaMA-Factory.
The configuration file (Config.yaml) is as follows:

\begin{figure}[H]
  \centering
  \begin{CodeBox}{Config.yaml for LoRA fine-tuning}
\noindent model\_name\_or\_path: 'Llama' or 'Qwen' \\
\noindent quantization\_bit: 4 \\
\noindent stage: 'sft' \\
\noindent do\_train: true \\
\noindent finetuning\_type: 'lora' \\
\noindent lora\_target: 'all' \\
\noindent lora\_rank: 16 \\
\noindent flash\_attn: 'fa2' \\

\noindent \#\#\# dataset \\
\noindent dataset\_dir: './data/' \\
\noindent dataset: 'POI' \\
\noindent template: 'llama3' \\
\noindent cutoff\_len: 8092 \\
\noindent max\_samples: null  \\
\noindent overwrite\_cache: true \\
\noindent preprocessing\_num\_workers: 384 \\

\noindent \#\#\# output \\
\noindent output\_dir: './output/' \\
\noindent logging\_steps: 200 \\
\noindent save\_steps: 1000 \\
\noindent plot\_loss: true \\
\noindent overwrite\_output\_dir: true \\

\noindent \#\#\# train
\noindent per\_device\_train\_batch\_size: 4 \\
\noindent gradient\_accumulation\_steps: 8 \\
\noindent learning\_rate: 1.0e-4 \\
\noindent num\_train\_epochs: 3.0 \\
\noindent lr\_scheduler\_type: 'cosine' \\
\noindent warmup\_ratio: 0.1 \\
\noindent weight\_decay: 0.05 \\

\noindent \#\#\# eval \\
\noindent val\_size: 0.1 \\
\noindent per\_device\_eval\_batch\_size: 1 \\
\noindent eval\_strategy: 'steps' \\
\noindent eval\_steps: 100 \\

\noindent \#\#\# hardware \\
\noindent gradient\_checkpointing: true \\
\noindent optim: 'adamw\_torch' \\
\noindent bf16: true

  \end{CodeBox}
  
\end{figure}

\section{Model Testing with RAG}
\label{app:RAG}

To simplify the deployment process of RAG, we leverage the Dify platform\footnote{\href{https://github.com/langgenius/dify}{https://github.com/langgenius/dify}} to streamline the RAG approach.
We divide it into two parts: knowledge base construction and model retrieval.

\subsection{Knowledge Base Construction}

In this work, we primarily store all POI information as content within the database.
We also construct a vector database according to all POIs that encapsulates the relevant details.
The embedding of this textual POI data is achieved using the bge-m3 model\footnote{\href{https://huggingface.co/BAAI/bge-m3}{https://huggingface.co/BAAI/bge-m3}}, which is specifically designed for efficient semantic embedding of text.

To manage the model's limitations on text length (\ie the maximum token limit), we break the content into segments of up to 4096 tokens.
To maintain semantic coherence between consecutive text segments, we introduce an overlap of 1024 tokens between adjacent segments.
This strategy ensures that the model retains context from one segment to the next, preserving the overall meaning even after segmentation.

Furthermore, to enhance the retrieval efficiency, we incorporate the deepseek-r1-32B model.
This model is used to generate a concise summary of each POI, producing up to 20 simple QA pairs.
These QA pairs help distill the most relevant information from the POI and improve the performance of the model during subsequent retrieval tasks by enabling faster matching.

When performing data retrieval, we leverage a pre-trained reranking model, bge-reranker-large\footnote{\href{https://huggingface.co/BAAI/bge-reranker-large}{https://huggingface.co/BAAI/bge-reranker-large}}. This reranker scores the vectorized text data, allowing us to rank the results and identify the top two most relevant POI entries. These top-ranked results are then returned to the user or further processed for the next stage in the retrieval pipeline.

This multi-step process ensures that the POI data is effectively stored, summarized, and retrieved, with an emphasis on semantic consistency and retrieval efficiency.

\subsection{Model Retrieval}

To simplify the retrieval process, we employ the visualization platform Dify to build a structured retrieval workflow, as shown in Figure~\ref{fig:app_retrieval_overview}. This workflow consists of five key components:

\begin{figure}
    \centering
    \includegraphics[width=0.85\linewidth]{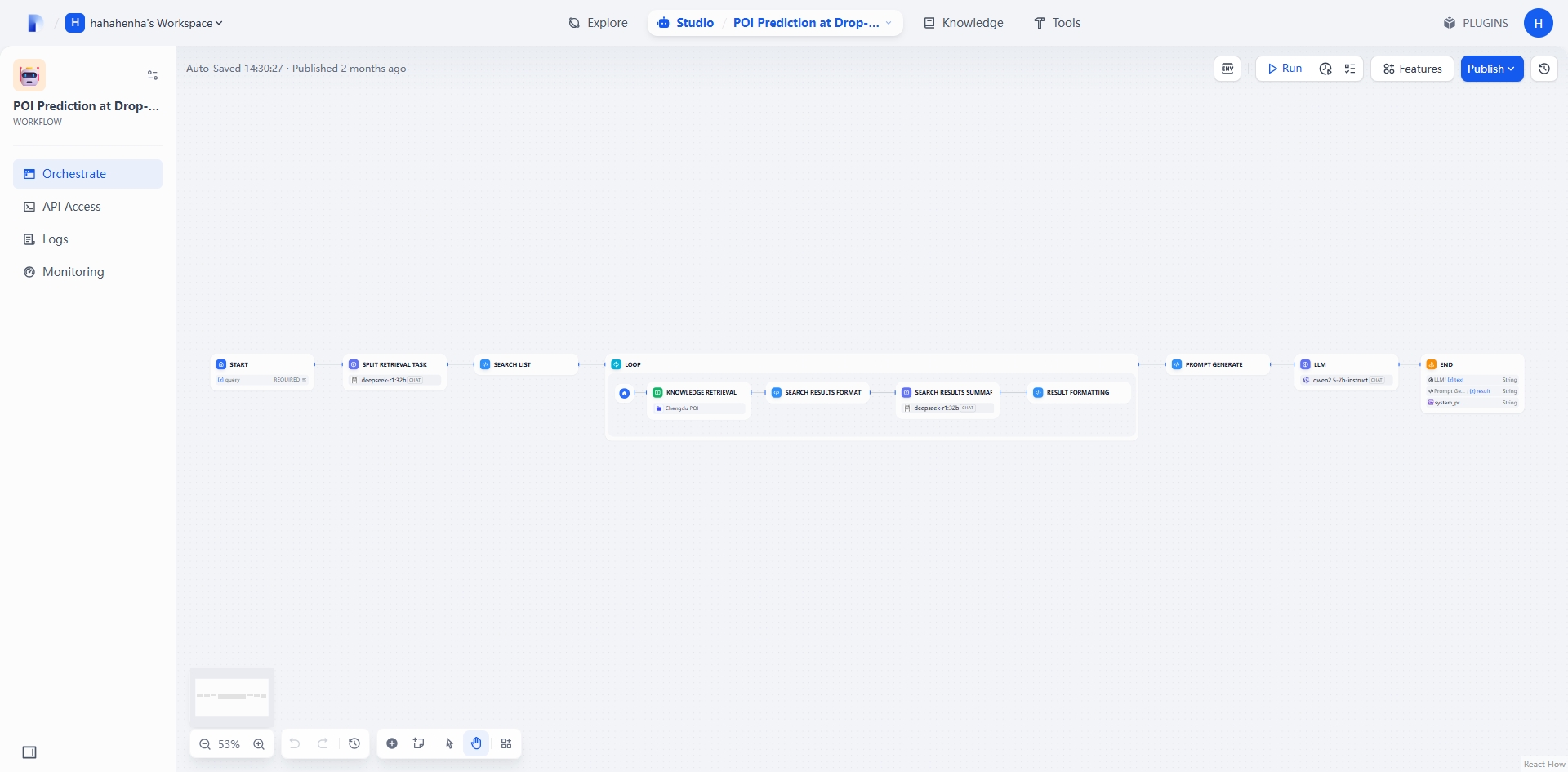}
    \caption{The overview of the retrieval process.}
    \label{fig:app_retrieval_overview}
\end{figure}

i) \textbf{Task Decomposition}: We utilize deepseek-r1-31B to break down the given query into manageable subtasks to enhance clarity and efficiency.
For each subtask, the deepseek propose a search term of no more than 15 characters for retrieval, as shown in Figure~\ref{fig:app_rag_stage1}.

\begin{figure}
\begin{CodeBox}{ Information Summarization}
{\bf Prompt:}\\
\# You want to search in order: RESEARCH TERM LIST\\

\# The following are the search results for RESEARCH TERM:\\
REARCH RESULTS\\
In the search results I gave you, each result is in the format of [content X begin]...[content X end], where X represents the numerical index of each article.\\
When answering, please pay attention to the following points:\\
- Not all the content of the search results is closely related to the user's question. You need to identify and filter the search results in combination with the question.\\
- For enumeration questions (such as enumerating all flight information), try to limit the answer to 10 points or less, and tell the user that they can check the search source and get complete information. Give priority to providing complete and most relevant enumeration items; if not necessary, do not actively tell the user content that is not provided by the search results.\\
- If the answer is long, please try to structure it and summarize it in paragraphs. If you need to answer in points, try to limit it to 5 points or less, and merge related content.\\
- You need to choose a suitable and beautiful answer format based on the user's requirements and the answer content to ensure strong readability.\\
- Your answer should be based on multiple relevant search results, and you cannot quote a web page repeatedly.\\
- Unless the user requires otherwise, the language of your answer should be consistent with the language of the user's question.\\

\# User message:\\
QUESTION\\

\# Please do not answer the user's question here, but summarize the key content, no more than 300 words.\\

\tcbline
\noindent\textbf{Example Question.}  

\# You want to search in order ....\\
\# The following are the search results for <yoo su zhuang mei yan zhong xin>:\\
<STYLESALON> distance: 130.8 meters\\
<SALON> distance: 110.32 meters\\
In the search results I gave you, ...\\

{\bf Expected Answer:}\\
The vehicle's destination is likely located in an area with various service amenities. Based on the starting points and the route taken, the nearby POIs are predicted to include:\\
1. \textbf{Beauty Salons}: The repeated presence of beauty salons suggests they are common in the area.\\
2. \textbf{Parking Lot}: Indicates a commercial or residential zone where parking is accessible.\\
3. \textbf{Laundry Shop}: Suggests proximity to other service facilities.\\
4. \textbf{Public Toilet}: Common in areas with high foot traffic or commercial activities.\\
5. \textbf{Dental Hospital}: Implies healthcare services are nearby.\\
These amenities collectively suggest the destination is in a mixed-use area with various conveniences for residents and visitors.
\end{CodeBox}
\caption{RAG stage ii.}
\label{fig:app_rag_stage1}
\end{figure}
\begin{figure}
    \centering
    \includegraphics[width=0.85\linewidth]{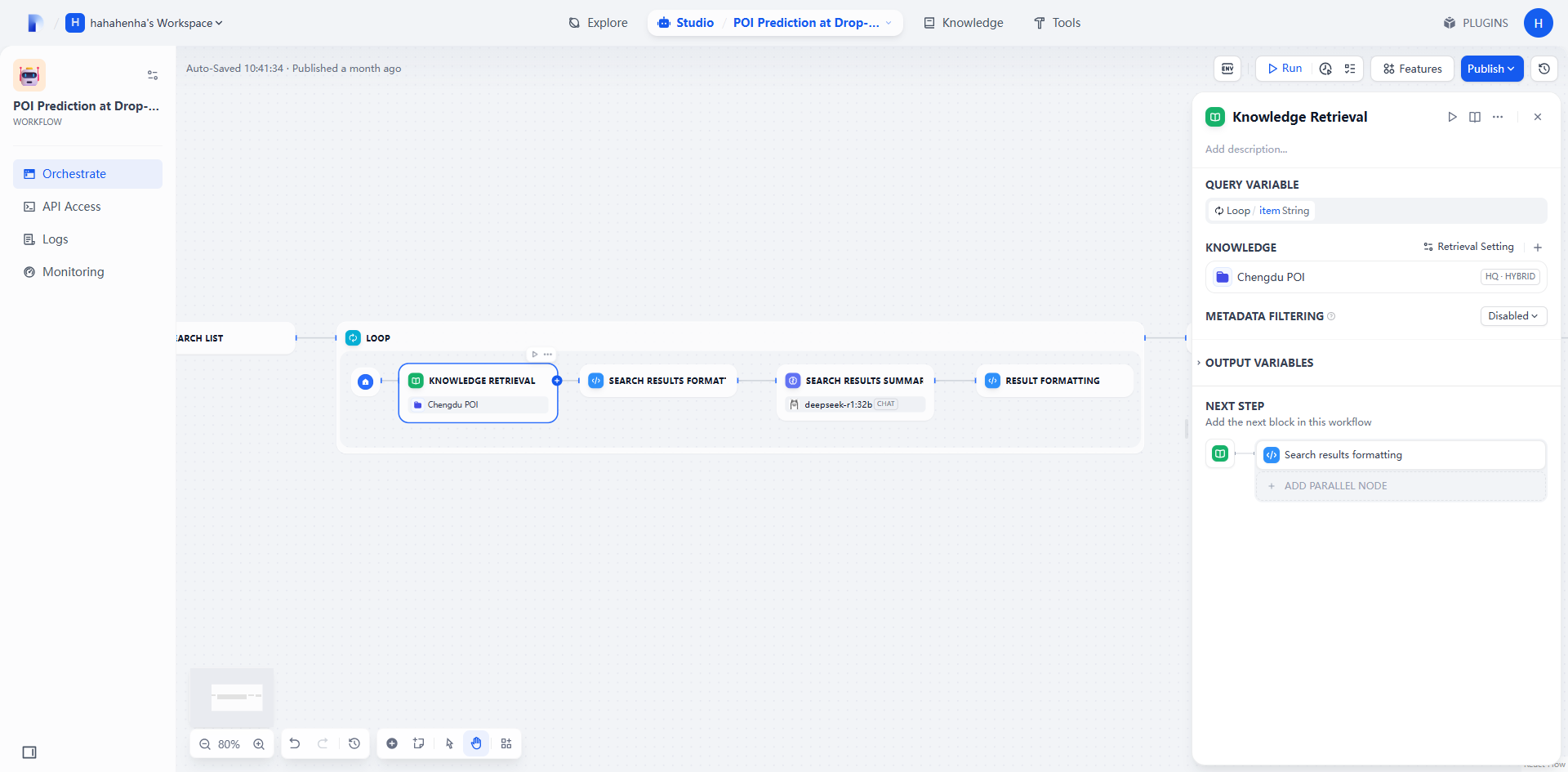}
    \caption{RAG stage iii.}
    \label{fig:app_dify_knowledge_retrieval}
\end{figure}


ii) \textbf{Extracting Search Terms}: We parse the formatted string returned by deepseek to identify and organize a list of relevant search terms to guide the following knowledge base query process.

iii) \textbf{Knowledge Base Searching}: We perform a systematic search for each identified search term in the knowledge base to collect comprehensive information. This step can be automatically completed in dify by simply configuring a knowledge base retrieval block and specifying the knowledge base we have built to be searched, as shown in Figure~\ref{fig:app_dify_knowledge_retrieval}.

iv) \textbf{Information Summarization}: We systematically search the knowledge base for each identified search term to collect comprehensive information. This step can be automatically completed in dify by simply configuring a knowledge base retrieval block and specifying the knowledge base we have built to be retrieved, as shown in Figure~\ref{fig:app_rag_stage4}.

\begin{figure}
\begin{CodeBox}{Task Decomposition}
{\bf Prompt:}\\
You are a useful intelligent assistant. You do not need to reply directly to the user's question, but break it down into a series of POI points that need to be searched in order (the name needs to be enclosed in angle brackets): <name> (large category, medium category, small category), and list them in the following format, so that you can provide users with more accurate answers based on the search information later:
[search text start]xxx[search text end],
[search text start]yyy[search text end],
...
Must be listed in the format of "[search text start]" and "[search text end]".\\

\tcbline
\noindent\textbf{Example Question.}  

Start Time: Tuesday, 10:43 am. Start POIs: [<er huan shuang lin bei zhi lu kou zhan (gong jiao zhan)>, <xu feng mei rong mei fa>, <yoo su zhuang mei yan zhong xin>,<jiu long cang yu yuan ting che chang>,<a ma ni xi hu hui suo>,<gong gong ce suo>,<cheng du rui an kou qiang>]. The vehicle passes the following POIs: [<er huan shuang lin bei zhi lu kou zhan (gong jiao zhan)>,<xu feng mei rong mei fa>,<jiu long cang yu yuan ting che chang>,<ya nao ke chuan qi>,<jin rui zhi leng>]. Please predict which POIs are near the vehicle's destination?\\

{\bf Expected Answer:}\\
\noindent[search text start]xufengmeirongmeifa[search text end],\\
\noindent[search text start]yoo su zhuang mei yan zhong xin[search text end],\\
\noindent[search text start]jiu long cang yu yuan ting che chang[search text end],\\
\noindent[search text start]a ma ni xi hu hui suo[search text end],\\
\noindent[search text start]gong gong ce suo[search text end],\\
\noindent[search text start]cheng du rui an kou qiang[search text end].
\end{CodeBox}
\caption{RAG stage iv.}
\label{fig:app_rag_stage4}
\end{figure}\begin{figure}
\begin{CodeBox}{Task Decomposition}
{\bf Prompt:}\\
You are a useful intelligent assistant. You do not need to reply directly to the user's question, but break it down into a series of POI points that need to be searched in order (the name needs to be enclosed in angle brackets): <name> (large category, medium category, small category), and list them in the following format, so that you can provide users with more accurate answers based on the search information later:
[search text start]xxx[search text end],
[search text start]yyy[search text end],
...
Must be listed in the format of "[search text start]" and "[search text end]".\\

\tcbline
\noindent\textbf{Example Question.}  

Start Time: Tuesday, 10:43 am. Start POIs: [<er huan shuang lin bei zhi lu kou zhan (gong jiao zhan)>, <xu feng mei rong mei fa>, <yoo su zhuang mei yan zhong xin>,<jiu long cang yu yuan ting che chang>,<a ma ni xi hu hui suo>,<gong gong ce suo>,<cheng du rui an kou qiang>]. The vehicle passes the following POIs: [<er huan shuang lin bei zhi lu kou zhan (gong jiao zhan)>,<xu feng mei rong mei fa>,<jiu long cang yu yuan ting che chang>,<ya nao ke chuan qi>,<jin rui zhi leng>]. Please predict which POIs are near the vehicle's destination?\\

{\bf Expected Answer:}\\
\noindent[search text start]xu feng mei rong mei fa[search text end],\\
\noindent[search text start]yoo su zhuang mei yan zhong xin[search text end],\\
\noindent[search text start]jiu long cang yu yuan ting che chang[search text end],\\
\noindent[search text start]a ma ni xi hu hui suo[search text end],\\
\noindent[search text start]gong gong ce suo[search text end],\\
\noindent[search text start]cheng du rui an kou qiang[search text end].
\end{CodeBox}
\caption{RAG stage iv.}
\label{fig:app_rag_stage4}
\end{figure}

v) \textbf{Contextual Prompt Construction}: Here, we use the summarized content as context to craft new prompts, enabling the model to generate accurate and relevant final results. The structure of the prompt is shown in Figure~\ref{fig:app_rag_stage5}.

\begin{figure}
\begin{CodeBox}{Final Prompt for RAG}
{\bf Prompt:}\\
\# User questions have the following format: The vehicle departure time is: xxx, and the starting point of the vehicle is near the following POIs: [xxx (large category, medium category, small category), ...] Please predict which POIs are near the vehicle's destination? Please reply to the user in the following format: The POIs near the vehicle's destination are: [xxx (major category, medium category, sub-category), ...].\\

\# User's question is: QUESTION \\

\# The search results are as follows: \\
\noindent[content 1 begin]CONTENT[content 1 end]\\
\noindent[content 2 begin]CONTENT[content 2 end]\\
...\\

In the search results I gave you, each result is in the format of [content X begin]...[content X end], where X represents the numerical index of each article.
When answering, please pay attention to the following points:

- Not all the content of the search results is closely related to the user's question. You need to identify and filter the search results in combination with the question.

- For enumeration questions (such as enumerating all flight information), try to limit the answer to 10 points or less, and tell the user that they can check the search source and get complete information. Give priority to providing complete and most relevant enumeration items; if not necessary, do not actively tell the user content that is not provided by the search results.

- If the answer is long, please try to structure it and summarize it in paragraphs. If you need to answer in points, try to limit it to 10 points or less, and merge related content.

- You need to choose a suitable and beautiful answer format based on the user's requirements and the answer content to ensure strong readability.

- Your answer should be based on multiple relevant search results, and you cannot quote a web page repeatedly.

- Unless the user requires otherwise, the language of your answer should be consistent with the language of the user's question.

\end{CodeBox}
\caption{RAG stage v.}
\label{fig:app_rag_stage5}
\end{figure}

This structured approach ensures that the retrieval process is both efficient and effective, leveraging Dify's capabilities to streamline operations.

\section{Detailed Evaluation Metrics}
\label{app:metrics}
We evaluate the performance of models on four answer types:
POI name, subcategory, medium category, and major category, covering spatiotemporal reasoning at multiple granularities. Specifically, we designed two evaluation settings differing in how the answer space is defined:

\paragraph{QA for Classification Tasks.}
For this finite-candidate task, every question is accompanied by an set of labels (categories).  
The model therefore returns a ranked list of at least $k$ candidates.  
We report Hit Ratio (HR@$k$) and Normalized Discounted Cumulative Gain (NDCG@$k$) at $k\!\in\!\{5,10,20\}$:
HR@$k$ measures the fraction of correct predictions within the top-k results, while NDCG@$k$ evaluates the ranking quality, emphasizing the positions of correctly predicted POIs.
Let $\hat{A} = \{\hat{a}_1, \hat{a}_2, \dots, \hat{a}_k\}$ be the model's top-k predictions and $A = \{a_1, a_2, \dots\}$ be the set of ground-truth labels, the metrics are defined as:
$$
\text{HR@}k = \frac{|\hat{A} \cap A|}{\min(k, |A|)}, \quad 
\text{NDCG@}k = \frac{1}{Z} \sum_{i=1}^{k} \frac{2^{rel(\hat{a}_i)} - 1}{\log_2(i+1)},
$$
where $rel(\hat{a}_i) = 1$ if $\hat{a}_i \in A$, and 0 otherwise.
The normalization term $Z$ represents the ideal DCG score under perfect ranking.

\paragraph{Open-world Generative QA.}
Here the model must generate free-form text containing at least $k$ POI names that satisfy the spatiotemporal constraints.  
After normalising the output format we still compute HR@$k$ and NDCG@$k$ (Full name matching) as above.  
In addition, to diagnose hallucination and lexical errors that HR/NDCG cannot capture, we calculate a textual-similarity score between the generated POI names and the ground truth:
$$
Similarity = \operatorname{BLEUScore}(\hat{A}, A)= BP \cdot \exp\left( \sum_{n=1}^{N}{w_n\cdot \log (p_n)} \right),
$$
where $\hat{A}$ is the predicted POI name, $A$ is the ground truth of the POI name, $p_n = \frac{\sum_{a \in S_{A}}{\mathrm{count}_n(a)}}{\sum_{a \in S_{\hat{A}}}{\mathrm{count}_n(a)}}$, $S_A$ is the n-gram set for the predicted POI name, $S_{\hat{A}}$ is the n-gram set for the ground truth of the POI name, $\mathrm{count}_n(a)$ Indicates the number of occurrences of n-gram $a$, $N= \max\{4, |A|, |\hat{A}| \}$, $|A|$ is the total length of ground truth, and $|\hat{A}|$ is the total length of the predicted POI name, $BP$ is the brevity penalty factor, which can be calculated by the following equation:
$$
BP= \begin{cases}1 & \text { if }|\hat{A}|>|A| \\ \exp \left(1-\frac{|A|}{|\hat{A}|}\right) & \text { if }|\hat{A}| \leq|A|\end{cases}.
$$

\section{Additional Experiments}

\subsection{Error Analysis}
\label{app:error_analysis}

\begin{figure}[t]
\centering
\subfigure[The number of records where the top-5 predictions are restaurants in different hours of the day]{
    \label{fig:error_analysis_a}
    \resizebox{0.475\linewidth}{!}{
        \begin{tikzpicture}
            \begin{axis}[
                ybar,
                tick align=inside, 
                grid=major,
        	xlabel=Hour of the day,
                ylabel=The number of records,
                ymin=0,
                ymax=8000,
                bar width=5pt,
                xmin=0,
                xmax=25,
                xtick={1,3,6,9,12,15,18,21,24},
                xticklabels={1,3,6,9,12,15,18,21,24},
                font=\normalsize
                ]
                \addplot [fill=mycolor55light, draw=mycolor55, postaction={pattern=crosshatch,pattern color=mycolor55}] table [x=x,y=y,,col sep=comma] {data/temporal_error.txt};
            \end{axis}
        \end{tikzpicture}
    }
}
\subfigure[Relationship between the farthest POI distance and the number of predicted records in the predicted POI list]{
\label{fig:error_analysis_b}
\resizebox{0.475\linewidth}{!}{
\begin{tikzpicture}
\begin{axis}[
    smooth,
    grid=major,
    xlabel=The farthest POI distance (Meter),
    xtick={0,1,2,3,4,5,6,7,8,9,10,11,12,13},
    xticklabels={,5,10,20,50,100,150,200,300,500,1000,1500,2000,>2000},
    xticklabel style={
        anchor=east,
        rotate=45,
        font=\normalsize
    },
    ylabel=The number of records,
    xmin=0,
    xmax=13,
    ymin=-10,
    font=\normalsize
    ]
    \addplot [each nth point=1, mark=*,mycolor55, line width=2pt] table [x=x, y=y,, col sep=comma] {data/spatial_error.txt};
\end{axis}
\end{tikzpicture}
}}

\subfigure[The number of predicted records v.s. the number of actual records]{
\label{fig:error_analysis_c}
\resizebox{0.475\linewidth}{!}{
\begin{tikzpicture}
\begin{axis}[
    ybar,
    tick align=inside, 
    grid=major,
    xlabel=The top 13 major categories,
    xtick={1,2,3,4,5,6,7,8,9,10,11,12,13,14,15,16,17,18,19},
    xticklabels={Catering Service,Companies and Enterprises,Transportation Facility Services,Lifestyle Services,Healthcare Services,Science, education, and cultural services,Government Agencies and Societies,Business Residence,Sports and Leisure Services,Accommodation Service,Financial Insurance Services,
    ,Public facilities,Scenic Spot,Automobile Service,Geographic Address Information,Car Sales,Motorcycle Service,Automobile Repair},
    xticklabel style={
        anchor=east,
        rotate=45,
        font=\normalsize
    },
    ylabel=The number of records,
    xmin=0,
    xmax=14,
    ymin=-10,
    bar width=3pt,
    ymax=10000,
    legend columns=1,
    legend style={font=\normalsize,at={(0.65,0.8)},anchor=south, text opacity=1, fill opacity=0,draw=none},
    font=\normalsize
    ]
    \addplot [fill=mycolor55light, draw=mycolor55, postaction={pattern=crosshatch,pattern color=mycolor55}] table [x=x, y=y1,, col sep=comma] {data/hallucinated_error.txt};
    \addplot [fill=mycolor42light, draw=mycolor42, postaction={pattern=bricks,pattern color=mycolor42}] table [x=x, y=y2,, col sep=comma] {data/hallucinated_error.txt};
    \legend{Actual number of records,
Predicted number of records}
\end{axis}
\end{tikzpicture}
}}
\subfigure[The impact of different prediction granularity on hit ratio]{
    \label{fig:error_analysis_d}
    \resizebox{0.475\linewidth}{!}{
    \begin{tikzpicture}
    \begin{axis}[
        ybar,
        tick align=inside, 
        grid=major,
    xlabel=Different granularities,
    xtick={1,2,3,4},
xticklabels={Major,Medium,Subcategory,\ \ \ \ \ \ \ \ \ \ \ \ \ \ \ \ Actual POI Name},
    xticklabel style={
        anchor=east,
        rotate=45,
        font=\normalsize
    },
        ylabel=Average HR@10,
        ymin=0,
        bar width=15pt,
        xmin=0,
        xmax=5,
        font=\normalsize
        ]
        \addplot [fill=mycolor55light, draw=mycolor55, postaction={pattern=crosshatch,pattern color=mycolor55}] table [x=x,y=y,,col sep=comma] {data/granularity_error.txt};
    \end{axis}
    \end{tikzpicture}
    }
}
\centering
\caption{Results for error analysis.}
\label{fig:error_analysis}
\end{figure}
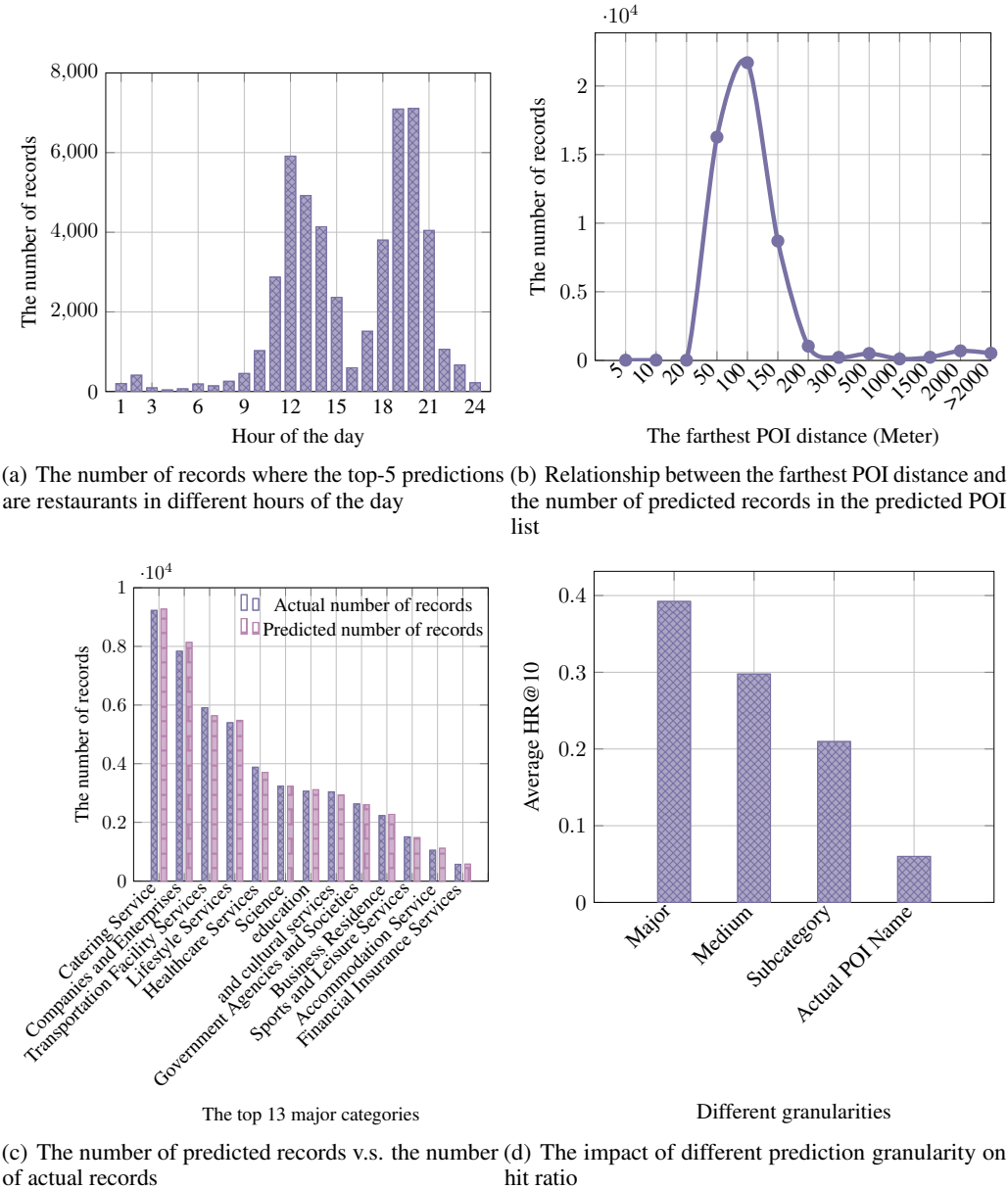

We manually inspected 50,000 representative errors produced by the best RAG+LoRA model across two tasks and grouped them into four categories, as demonstrated in Figure~\ref{fig:error_analysis}:

\begin{enumerate}[leftmargin=1.4em,label=\arabic*)]
    \item \textbf{Temporal misalignment}: As shown in Figure~\ref{fig:error_analysis_a}, we visualized the number of passenger records indicating visits to restaurants throughout the day. While there is a clear trend of increasing predicted records during meal times, we also observe some abnormal predictions, such as a small number of results suggesting passengers visited restaurants between 3 and 5 a.m. This highlights temporal misalignment as a notable challenge in working with \name\ dataset.
    \item \textbf{Spatial over-reach}:
          For each prediction result, we calculated the distance between the two farthest POIs in the list of nearbyPOIs at the destination, the relationship between this distance and the number of record is shown in Figure~\ref{fig:error_analysis_b}. We found that some predictions still exhibited notable anomalies, such as records where predicted POIs were more than 1 km apart. This demonstrates that current advanced LLMs have limitations in processing geospatial information, underscoring another critical challenge when analyzing spatiotemporal data.
    \item \textbf{Hallucinated entity}:
          For QA tasks, hallucination poses another significant obstacle to model accuracy. In the \name dataset, we analyzed the number of unique POI names across the first 13 major categories and compared them with those generated by predictions, as shown in Figure~\ref{fig:error_analysis_c}. Notably, within the ``Companies and Enterprises'' category, the model fabricated a substantial number of POIs that were absent from the original dataset. This highlights the need for improved reliability in generating accurate and dataset-consistent responses.
    \item \textbf{Granularity confusion}:
          As shown in Figure~\ref{fig:error_analysis_d}, the model has higher performance for correctly predicting major or medium category but not subcategory or specific POI name.
\end{enumerate}

These cases are prevalent in the experiments, which poses great challenges for the tasks. To further enhance the performance on our dataset, it is crucial to consider better methods to improve the spatiotemporal reasoning capabilities.

\end{document}